\documentclass[sigconf]{acmart}

\AtBeginDocument{%
  }

\usepackage{amsmath}
\usepackage{graphicx} 
\usepackage{microtype}
\usepackage{booktabs} 
\usepackage{multirow}
\usepackage{graphicx} 
\usepackage{float}
\usepackage{CJKutf8}
\usepackage{bookmark}
\usepackage{amsfonts}
\usepackage{amsmath}
\usepackage{adjustbox}
\usepackage[ruled,linesnumbered]{algorithm2e}
\usepackage{marvosym}
\copyrightyear{2024}
\acmYear{2024}
\setcopyright{rightsretained}
\acmConference[KDDCup '24]{To be presented at KDD Cup 2024 Workshop:
More diverse more adaptive: Comprehensive Multi-task Learning for Improved LLM Domain Adaptation in E-commerce}{August 25, 2024}{Barcelona, Spain}
\acmBooktitle{KDD Cup 2024 Workshop: A Multi-task Online Shopping Challenge for Large Language Models, August 25, 2024, Barcelona, Spain}
\acmDOI{}
\acmISBN{}

\begin{document}
\begin{CJK}{UTF8}{gbsn}

\title{More diverse more adaptive: Comprehensive Multi-task Learning for Improved LLM Domain Adaptation in E-commerce}

\author{Piao Tong$^{*}$}
\email{piaot@std.uestc.edu.cn}
\affiliation{%
  \institution{University of Electronic Science \\and Technology of China}
  \city{Chengdu}
  \country{China}
}

\author{Pei Tang$^{*}$}
\email{richpp523@gmail.com}
\affiliation{%
  \institution{Xiaohongshu Inc.}
  \city{Shanghai}
  \country{China}
}

\author{Zhipeng Zhang$^{*}$}
\email{zhangzhipeng.work@bytedance.com}
\affiliation{%
  \institution{ByteDance Inc.}
  \city{Shanghai}
  \country{China}
}

\author{Jiaqi Li}
\email{slrrr8848@gmail.com}
\affiliation{%
  \institution{Sichuan University}
  \city{Chengdu}
  \country{China}
}

\author{Qiao Liu}
\email{qliu@uestc.edu.cn}
\affiliation{%
  \institution{University of Electronic Science \\and Technology of China}
  \city{Chengdu}
  \country{China}
}

\author{Zufeng Wu\textsuperscript{\Letter}}
\email{wuzufeng@uestc.edu.cn}
\affiliation{%
  \institution{University of Electronic Science \\and Technology of China}
  \city{Chengdu}
  \country{China}
}

\renewcommand{\authors}{Zhipeng Zhang, Piao Tong, Qiao Liu, Yingwei Ma∗, Xujiang Liu, Xu Luo}
\renewcommand{\shortauthors}{Zhipeng and Piao, et al.}




\begin{abstract}
In recent years, Large Language Models (LLMs) have been widely applied across various domains due to their powerful domain adaptation capabilities. Previous studies have suggested that diverse, multi-modal data can enhance LLMs' domain adaptation performance. However, this hypothesis remains insufficiently validated in the e-commerce sector. To address this gap, we propose a comprehensive e-commerce multi-task framework and design empirical experiments to examine the impact of diverse data and tasks on LLMs from two perspectives: "capability comprehensiveness" and "task comprehensiveness." Specifically, we observe significant improvements in LLM performance by progressively introducing tasks related to new major capability areas and by continuously adding subtasks within different major capability domains. Furthermore, we observe that increasing model capacity amplifies the benefits of diversity, suggesting a synergistic relationship between model capacity and data diversity.
Finally, we validate the best-performing model from our empirical experiments in the KDD Cup 2024, achieving a rank 5 in Task 1. This outcome demonstrates the significance of our research for advancing LLMs in the e-commerce domain.
\let\thefootnote\relax\footnotetext{*~~~The first four authors contributed equally to this work.}
\let\thefootnote\relax\footnotetext{\Letter~~Corresponding Author.}
\end{abstract}



\maketitle

\section{Introduction}

Recently, Large Language Models (LLMs) have been deployed across various domains\cite{zhang2023language, li2023text}, leveraging their robust domain adaptation capabilities and the flexibility of generative models, which are not constrained by the task-specific fine-tuning requirements of discriminative models. This adaptability allows LLMs to easily transfer to different fields and achieve impressive performance.

The e-commerce sector is rich in textual data, including product titles, descriptions, reviews, and Q\&A content\cite{tong2024care, tong2024interpretable}. Researchers have begun to utilize these data sources as inputs for LLMs in e-commerce applications, yielding promising results. These studies encompass a wide range of tasks, from classic NLP extraction and classification to product sequence recommendation and query ranking. However, most research focuses on Prompt Engineering or fine-tuning for specific tasks. This trend raises an intriguing and pressing research question: How do large models perform in multi-task settings, and what is the optimal approach to constructing a multi-task framework for the e-commerce domain?

\begin{figure*}[htbp]
  \centering 
  \includegraphics[width=0.7\textwidth]{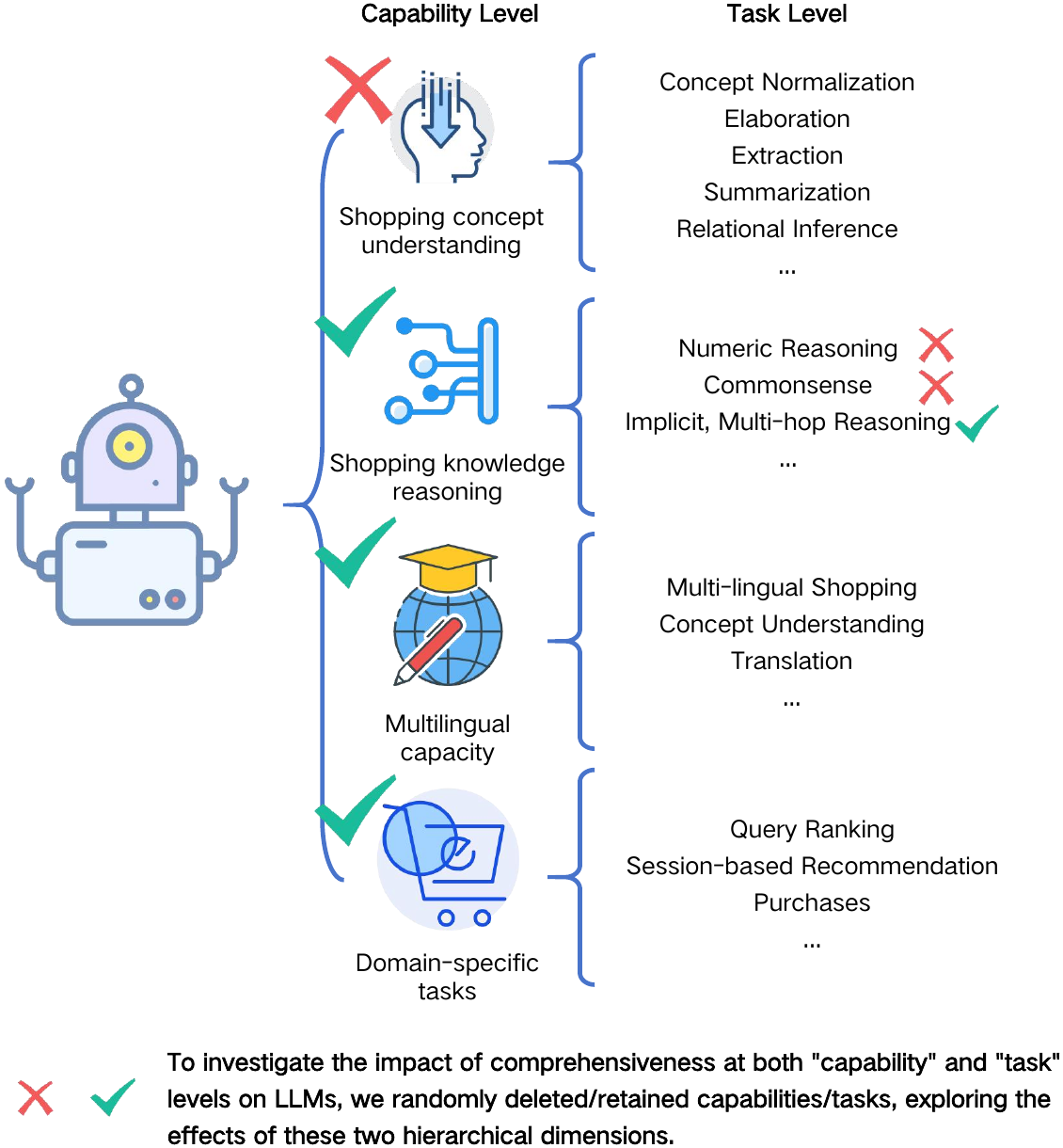}
  \caption{Our Experimental Framework: Ablation Studies at Capability and Task Levels to Evaluate the Impact of Data Diversity on Large Model Performance.}
  \label{fig:figure_1}\vspace{-1em}
\end{figure*}
Early methodologies, such as P5\cite{geng2022recommendation}, were proposed to introduce a wider range of tasks to explore model performance across various recommendation system-related scenarios. However, their multi-task design remained confined to the realm of recommendations. Subsequently, researchers attempted to collate open-source data from diverse e-commerce domains\cite{peng2024ecellm}, manually categorizing task types to construct an ontological structure. This led to the delineation of categories such as "Product Understanding," "User Understanding," "Query Product Matching," and "Product QA." Nevertheless, the framework's integrated data lacked comprehensiveness, primarily due to its insufficiently high-level and generalizable task classification (e.g., inability to categorize reasoning tasks and multilingual tasks). The ShopBench framework, introduced in KDD CUP 2024, addressed these limitations. It constructed a comprehensive multi-task learning framework for large language models based on the sequence of human cognitive patterns, progressing from "common sense cognition to logical reasoning, followed by fine-tuning on downstream tasks." However, this framework also gave rise to new research questions, particularly regarding how the capabilities of large language models evolve with respect to both the comprehensiveness of the framework and the diversity of tasks.

In this paper, we address the capabilities integrated into large language models within the ShopBench framework by designing a diverse array of tasks that align with e-commerce business logic. These tasks are highly heterogeneous and numerous, designed to test various aspects of model performance. We employ multiple sensible approaches to acquire task-specific data, including prompt engineering and open-source data collection. Notably, we conducted extensive empirical experiments based on these data, aiming to explore the impact of diverse data and tasks on LLMs from two perspectives: "capability comprehensiveness" and "task comprehensiveness." Specifically, we investigated these perspectives by ablating major capability domains of the large language model and controlling the number of subtasks corresponding to each major capability domain. This research provides valuable insights into comprehensively constructing e-commerce multi-task learning frameworks for large language models and expanding the capability boundaries of e-commerce LLMs. Additionally, we posit that as data diversity increases, the fine-tuning capacity of large language models must be concurrently enhanced to accommodate more diverse features. We substantiate this hypothesis by manipulating capacity parameters.

\section{METHODOLOGY}
\subsection{Experimental Framework}
We first designed corresponding subtasks for the four key capabilities of large language models mentioned in ShopBench. We then conducted ablation experiments at both "Capability Level" and "Task Level" to evaluate the impact of "Capability comprehensiveness" and "Task diversity" on LLM domain adaptation performance. When designing tasks, two key points were considered:
\begin{figure*}[htbp]
  \centering 
  \includegraphics[width=1\textwidth]{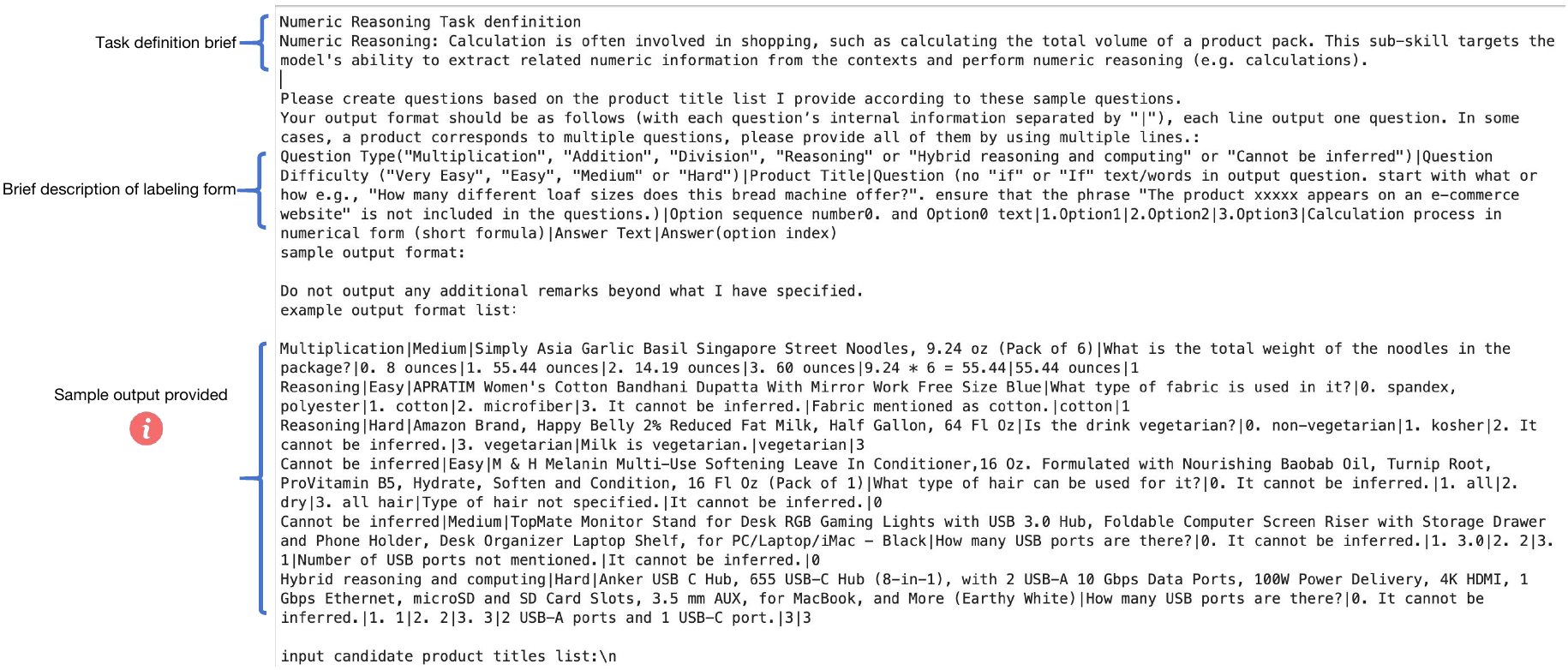}
  \caption{Exemplar Instructions for GPT-4 in Generating Training Data (Reasoning Tasks).}
  \label{fig:figure_2}\vspace{-1em}
\end{figure*}
\begin{itemize}
\item Task heterogeneity: Even subtasks within a single Capability should be as diverse as possible.
\item Alignment with e-commerce business logic: Tasks unrelated to the e-commerce domain may potentially hinder LLM performance in e-commerce applications.
 \end{itemize}
 
\subsection{Data Acquisition for Training}

For each task we designed, corresponding training data must be acquired. We obtain data through two approaches: (1) Collecting open-source data, and (2) Generating training data using high-performance LLMs like GPT-4.

Important tip: When aiming to make LLM-generated annotations possess certain characteristics, relying solely on prompt engineering may be insufficient. For instance, to include challenging samples in the output, an incorrect and correct approach are as follows:
\begin{itemize}
\item Incorrect: Including "Please ensure your annotations contain some difficult questions to thoroughly train my model" in the instruction.

\item Correct: Explicitly incorporating desired output sample characteristics using delimiters in the "Sample output" provided to the LLM. As shown in Figure 3, we include "|Question Difficulty ("Very Easy", "Easy", "Medium" or "Hard")|" in the "Sample output" to prompt the model to consider question difficulty when generating the dataset.
\end{itemize}

\subsection{LoRA Rank Control During Fine-tuning}

A logical approach suggests that as data diversity increases, the parameter volume during model fine-tuning should correspondingly increase to accommodate the data's diversity. Currently, LoRA is the predominant fine-tuning method, utilizing two low-rank matrices to generate matrices of the same size as LLM components, which are then added to the original parameter matrices. By controlling the rank of these low-rank matrices, we can regulate the number of fine-tuning parameters. While previous studies have examined the relationship between parameter quantity and LLM performance, few have investigated how rank changes affect LLM performance under varying data diversity conditions. Therefore, we attempt to control both diversity and rank to explore the relationship between fine-tuning parameter capacity and data diversity.

\section{Experiments}
\subsection{Overall Performance}
Our approach achieved considerably performance gain over the baseline solution, and ranked top-5 in task-1.
The main results are shown in Table~\ref{tab:table_1}. 

\begin{table}[h]

  \centering \begin{tabular}{lcc}
\specialrule{0.7pt}{0pt}{2pt}
\textbf{Dataset} & \multicolumn{1}{l}{\textbf{Metric}} & \multicolumn{1}{l}{\textbf{Ranking}} \\ \specialrule{0.4pt}{0pt}{2pt}
track1        & Score=0.803                      & \textbf{5th}                                  \\
track4       & Score=0.707                      & 7th                     \\
track5       & Score=0.745                      & 6th                     \\
\specialrule{0.7pt}{1pt}{2pt}
\end{tabular}
  \caption{Performance of our approach.}
  \label{tab:table_1} 
  \vspace{-3.0em}
\end{table}

\subsection{Ablation Studies}
We conduct ablation studies on the training data from two perspectives: "capability comprehensiveness" and "task comprehensiveness," while simultaneously controlling the rank of LoRA to investigate the synergistic effects of model parameter capacity and data diversity on LLM performance. Our experiments yield two key findings:
\begin{table*}[htbp]
\centering
\begin{tabular}{lcc}
\specialrule{0.7pt}{1pt}{2pt} 
\multicolumn{1}{c}{\textbf{Variants}}                                & \textbf{Track1 Score} \\
\specialrule{0.4pt}{0pt}{2pt} 
Only generating and training Task3 data & 0.692              \\
Preliminary addition of Track1 data & 0.745              \\
Reducing data from 30,000 to 2,000 samples & 0.758              \\
Adding tasks related to Track3 and Track2 & 0.774              \\
Adjusting from 2,000 to 6,000 samples & 0.784              \\
Increasing LoRA rank from 16 to 32 & \textbf{0.803}              \\

\specialrule{0.7pt}{1pt}{2pt} 
\end{tabular}
\caption{Ablation Study on Model Selection, Training Data Choice, and LoRA Rank.}
\label{tab:table_2}
\vspace{-2.0em}
\end{table*}

\begin{itemize}
\item When developing LLMs, incorporating a more comprehensive and diverse set of capabilities through varied data leads to qualitative improvements in model performance. Conversely, blindly increasing data volume with homogeneous content can actually degrade model effectiveness.
\item As data diversity increases, it is crucial to concurrently expand the parameter capacity during LLM fine-tuning, enabling the model to effectively accommodate and leverage the enhanced data diversity.
\end{itemize}
\section{Conclusion and Future Work}
This study, through empirically designed experiments, underscores the critical importance of comprehensive and diverse data in developing effective LLMs. Furthermore, the synergistic relationship between data diversity and LoRA rank variations in enhancing LLM performance demonstrates that increased parameter capacity is necessary to accommodate diverse information within the data. In future work, we plan to explore additional methods for expanding model capacity, such as MMoE and multi-LoRA, and further investigate their synergistic effects with data diversity in advancing LLMs within the e-commerce domain
\vspace{-0.3em}
\bibliographystyle{ACM-Reference-Format}
\bibliography{custom-base-leaderboardcar}

\end{CJK}
\end{document}